\documentclass[letterpaper, 10 pt, conference]{ieeeconf}
\IEEEoverridecommandlockouts
\usepackage{amsmath,amssymb,amsfonts}
\DeclareMathOperator*{\argmax}{arg\,max}

\newcommand{\pluseq}{\mathrel{+}=}

\usepackage{graphicx}
\usepackage{textcomp}
\usepackage{xcolor}
\def\BibTeX{{\rm B\kern-.05em{\sc i\kern-.025em b}\kern-.08em
		T\kern-.1667em\lower.7ex\hbox{E}\kern-.125emX}}

\usepackage{tikz}
\usetikzlibrary{shapes,arrows,positioning}

\usepackage{mathtools}
\usepackage{bm}
\newcommand\norm[1]{\left\lVert#1\right\rVert}
\usepackage{empheq}

\usepackage{neuralnetwork}
\newcommand{\h}[2]{$L$}
\newcommand{\mf}[2]{$F_1$}
\newcommand{\ms}[2]{$F_2$}

\usepackage[font=footnotesize]{caption}
\usepackage{subcaption}

\usepackage[utf8]{inputenc}

\usepackage{svg}
\graphicspath{ {./pics/} }

\usepackage{multirow}

\usepackage[sorting=none]{biblatex}
\usepackage{hyperref}

\tikzstyle{block} = [draw, rectangle, 
minimum height=0.8cm, minimum width=1.6cm]
\tikzstyle{vblock} = [draw, rectangle, 
minimum height=0.8cm, minimum width=1.6cm, node distance=1cm]
\tikzstyle{sum} = [draw, circle, node distance=2cm]
\tikzstyle{knot} = [coordinate]
\tikzstyle{sknot} = [coordinate, node distance=1cm]
\tikzstyle{input} = [coordinate]
\tikzstyle{output} = [coordinate]
\tikzstyle{soutput} = [coordinate, node distance=1cm]
\tikzstyle{pinstyle} = [pin edge={to-,thin,black}]

\title{Autonomous Blimp Control via H$_{\infty}$ Robust  Deep Residual Reinforcement Learning}

\author{Yang Zuo$^{2,*}$,  Yu Tang Liu$^{1,2,*}$, Aamir Ahmad$^{1,2}$
	\thanks{$^1$Max Planck Institute for Intelligent Systems, 72076 T{\"u}bingen, Germany.
		$^2$Institute of Flight Mechanics and Controls, University of Stuttgart, 70569 Stuttgart, Germany.{ zuoyang0601@gmail.com, yutang.liu@tuebingne.mpg.de, aamir.ahmad@ifr.uni-stuttgart.de}
		$^*$Yang Zuo and Yu-Tang Liu contribute equally to this work as the first author. }
}

\begin{document}
	\maketitle
	
	\begin{abstract}
		Due to their superior energy efficiency, blimps may replace quadcopters for 
		long-duration aerial tasks. However, designing a controller for blimps to 
		handle complex dynamics, modeling errors, and disturbances remains an unsolved 
		challenge. One recent work combines reinforcement learning (RL) and a PID 
		controller to address this challenge and demonstrates its effectiveness in 
		real-world experiments. In the current work, we build on that 
		using an $H_{\infty}$ robust controller to expand the stability margin and 
		improve the RL agent's performance. Empirical analysis of different mixing 
		methods reveals that the resulting H$_{\infty}$-RL controller outperforms the 
		prior PID-RL combination and can handle more complex tasks involving intensive 
		thrust vectoring. We provide our code as open-source at 
		\url{https://github.com/robot-perception-group/robust_deep_residual_blimp}. 
	\end{abstract}

	\section{Introduction}
	Unmanned aerial vehicles (UAVs) like multirotors and fixed 
	wings are increasingly being used for visual tracking tasks such as aerial 
	cinematography, wildlife monitoring\cite{Gonzalez2016}, and 
	precision farming\cite{duggal2016plantation}. However, while multirotors have 
	limitations such as short battery life and small payload, fixed-wings 
	must constantly move to stay airborne. We propose using autonomous blimps, which 
	are more energy-efficient and have a higher payload for long-duration, 
	small-region hovering tasks.
	
	Blimp control, however, presents challenges in the context of modeling 
	uncertainties and wind disturbances. Prior work used a deep residual 
	reinforcement learning (DRRL) framework\cite{silver2018residual, 8794127} to 
	address this with a model-free proportional-integral-derivative (PID) base 
	controller and an RL agent \cite{Liu.20220310}. During training, the RL agent's 
	action can be considered as an extra disturbance to the base controller, so the 
	robustness of the base controller defines the permitted exploratory actions.
	
	In the current work, we replace the PID base controller with a robust 
	model-based $H_{\infty}$ controller to expand the stability margin. The 
	$H_{\infty}$ robust design framework generates a controller that makes decisions 
	based on the worst-case scenario, which offers the most significant safety bound 
	at the cost of control performance. This gives the RL agent a larger exploration 
	bandwidth and more potential performance growth. The model-based approach also 
	allows deriving the worst-case bound that considers the total amount of model 
	uncertainty and disturbance from both the environment and the RL agent.
	
	We show in the simulated environment that the DRRL agent, consisting of the $H_{\infty}$ robust control and a proximal policy optimization (PPO) agent\cite{schulman2017proximal}, outperforms 
	the previous PID-PPO combination in performance and robustness and can even 
	handle more challenging tasks. We also improve the DRRL framework by a variable mixing factor such that the controller can grant the RL agent a variable amount 
	of control authority. We designed the base controller's thrust vectoring to 
	enhance the final performance further, allowing the RL agent to access a 
	more significant state and action space for better exploration.
	
	\begin{figure}[t]
		\centering
		\begin{subfigure}[t]{0.45\textwidth}
			\centering
			\includegraphics[width=1.0\linewidth, trim={0cm 5cm 5cm 0cm},clip]{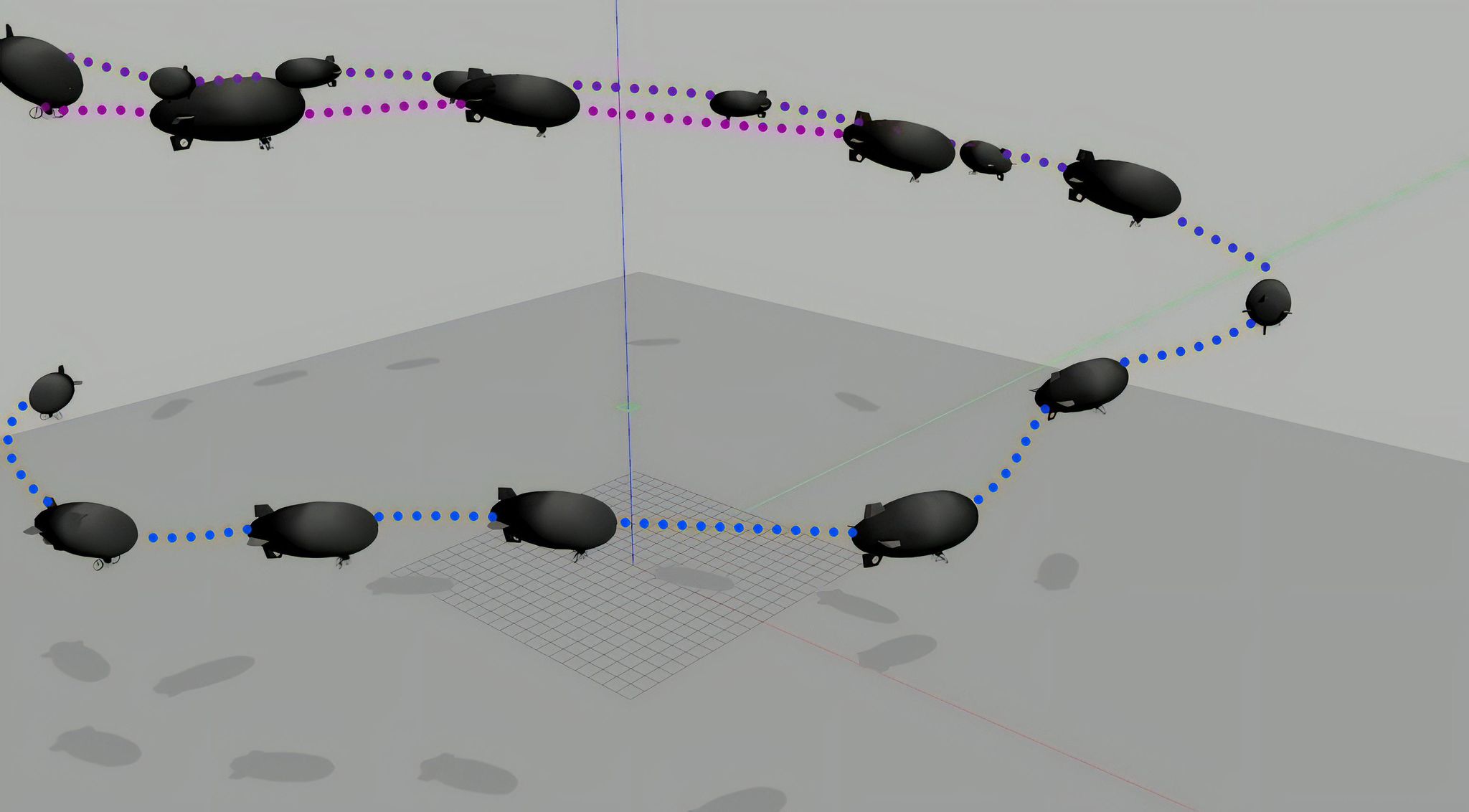}
		\end{subfigure}
		~
		\begin{subfigure}[t]{0.45\textwidth}
			\centering
			\includegraphics[width=1.0\linewidth, trim={2cm 4cm 4cm 2cm},clip]{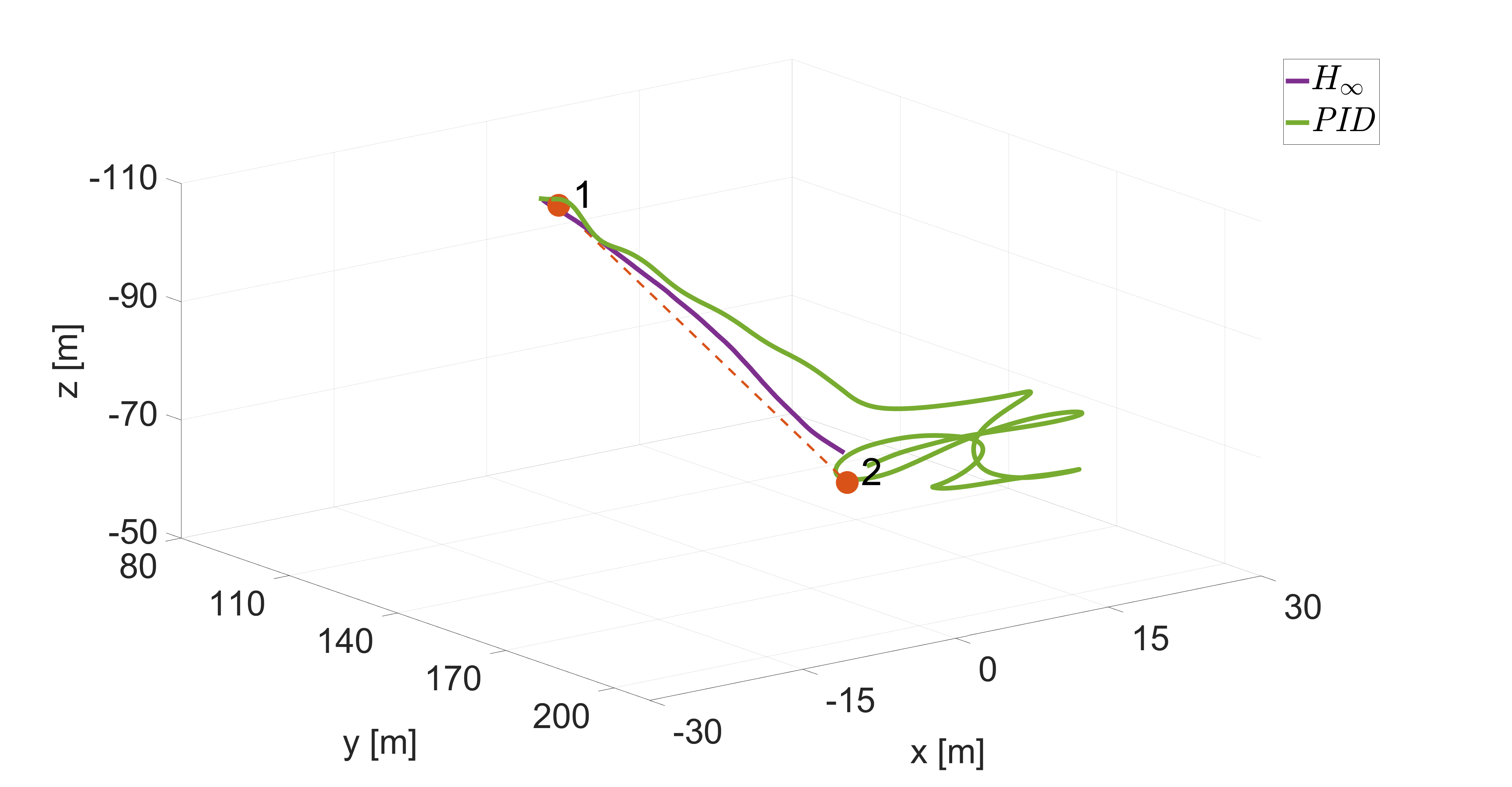}
		\end{subfigure}
		\caption{Top: the simulated blimp with the proposed $H_{\infty}$-PPO controller in the challenging coil trajectory. Bottom: descend trajectory of our $H_{\infty}$-PPO versus prior PID-PPO \cite{Liu.20220310} controller. Our controller is more robust against disturbance and improves altitude control by utilizing thrust vectoring, while PID-PPO controller can only rely on the elevators for altitude control. As a result, it can deviate nearly 15 meters from the desired path.}
		\label{fig: cover}
	\end{figure}

	\section{Related Work}
	Research on reliable robotic platforms for aerial tracking tasks has led to the 
	exploration of blimp and vision-based control, with most using PID-based 
	control \cite{price2022perception,  zhang1999visual, 1626776, 
		price2022simulation, Takaya2006PIDLO}. However, PID controllers are often a 
	suboptimal solution for non-linear control problems like a blimp. Alternative 
	solutions from model-based control frameworks, such as optimal control 
	\cite{fukushima2006model, 1013654}, adaptive control 
	\cite{liu2020adaptive}, or robust control \cite{cheng2019robust} have been 
	sought, but these have not yielded reliable controllers for real-world 
	experiments due to model uncertainty and output disturbance. In recent years, 
	RL with Gaussian Process-based models has been used for low-dimensional tasks 
	\cite{rottmann2009adaptive, 4399531, ko2007gaussian}. In contrast, Deep RL 
	(DRL) with large capacity models achieved a 3D path-following task in 
	simulation \cite{nie2019three} and in the first real-world experiment with the 
	data-driven approach \cite{Liu.20220310}. This work has extended the prior DRRL 
	agent \cite{Liu.20220310} by replacing PID with a robust $H_{\infty}$ controller 
	to improve safety and performance growth.

	\section{Methodology}
	In this section, we first introduce the simulators and formulate the task in the reinforcement learning framework (Sec.~\ref{sec: task formulation}). Different from \cite{Liu.20220310}, we introduce the $H_{\infty}$ controller as our base control (Sec.~\ref{sec: hinf robust control}). Lastly, we introduce our robust $H_{\infty}$-based deep residual reinforcement learning controller (Sec.~\ref{sec: robust hinf-rl in bimpsim}) as shown in Fig.~\ref{mixture}, where mixer block represents the following equation, 
	\begin{equation}
		a_{mixed} = (1-q)a + qu \label{eqn: mixture}
	\end{equation}

	\tikzstyle{block} = [draw, rectangle, 
	minimum height=0.4cm, minimum width=0.8cm]
	\tikzstyle{vblock} = [draw, rectangle, 
	minimum height=0.4cm, minimum width=0.8cm, node distance=0.7cm]
	\tikzstyle{hblock} = [draw, rectangle, 
	minimum height=0.4cm, minimum width=0.8cm, node distance=3.2cm]
	\tikzstyle{sum} = [draw, circle, node distance=1cm]
	\tikzstyle{knot} = [coordinate]
	\tikzstyle{sknot} = [coordinate, node distance=0.6cm]
	\tikzstyle{input} = [coordinate]
	\tikzstyle{output} = [coordinate]
	\tikzstyle{soutput} = [coordinate, node distance=0.5cm]
	\tikzstyle{pinstyle} = [pin edge={to-,thin,black}]
	\begin{figure}[h]
		\centering
		\begin{tikzpicture}[auto, node distance=2.6cm,>=latex']		
			\node [input, name=input] {};
			\node [knot, right of=input] (node1) {};
			\node [block, right of=node1] (agent) {$a_t\sim\pi(\cdot|s_t)$};
			\node [knot, right of=agent] (node2) {};
			\node [output, right of=node2] (output){};
			\node [block, below of=agent, node distance=1.2cm] (k){$u_t=H_{\infty}(s_t)$};
			\node [hblock, right of=k] (m){Mixer (\ref{eqn: mixture})};
			\node [block, below of=k, node distance=1cm] (env){Environment};
			\node [knot, left of=k] (node3) {};
			\draw [->] (agent) -| node[pos=0.8]{$a_t$}(m);
			\draw [->] (m) |- node[pos=0.4]{$a_{mixed,t}$}(env);
			\draw [->] (env) -- (node1 |- env) |- node[pos=0.35]{$\;s_{t+1}, r_t$}(agent);
			\draw [->] (node3) -- node[pos=0.5]{$s_{t+1}$}(k);
			\draw [->] (k) -- node[pos=0.4]{$u_{t}, q_{t}$} (agent);
			\draw [->] (k) -- node[pos=0.5]{$u_{t}, q_{t}$} (m);
		\end{tikzpicture}
		\caption{Our robust deep residual reinforcement learning framework. Every time step, the mixer gathers the action command from the policy $a_t$ and the controller $u_t$ and then mixes them based on the mixing factor $q_t$ evaluated by the controller.}
		\label{mixture}
	\end{figure}
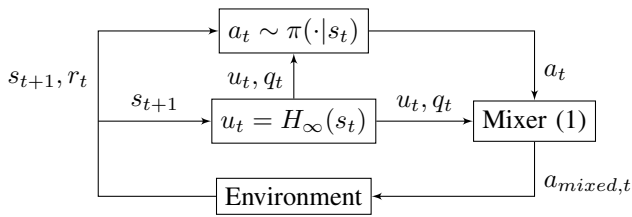
	
	The second difference from the previous work \cite{Liu.20220310}, which applies a fixed number of $q$, is that we sample it randomly from a distribution. The variable $q$ allows the controller to decide how much authority can be granted to the RL agent, depending on the situation. For example, when the wind disturbance is prominent, the controller can increase $q$ for more intervention and safety.  
	
	In the experiments section, we demonstrate that reducing the amount of intervention from the base control $q$ improves the final performance (Sec.\ref{sec: experiment turtle}). Therefore, our goal is to design a robust controller to guarantee control stability during both learning and testing phase while a minimum amount of intervention is required.

	\subsection{Markov Decision Process (MDP)}
	We first formulate RL problems as an MDP, and it can be represented as a tuple, $\mathcal{M}=(\mathcal{S}, \mathcal{A}, R, \mathcal{P}, \gamma, \rho)$, where $s_t \in \mathcal{S}$ and $a_t \in \mathcal{A}$ are the state and action space respectively. At any time step $t\in\mathbb{R}$, the
	RL agent samples an action from its control policy based on the observed environmental state $a_t\sim\pi(\cdot|s_t)$. Then the environment returns the next state and a reward base on the underlying transition dynamics $s_{t+1}\sim\mathcal{P}(\cdot|s_t,a_t)$ and a reward function $r_t=R(s_t,a_t)$, which defines the desired behavior and can be viewed as a task description. 
	Given the discount factor $\gamma \in [0,1)$ and initial state distribution $s_0\sim \rho(\cdot)$, the goal of the RL agent is to find a control policy such that the total amount of discounted reward can be maximized, i.e. 
	\begin{align}
		\pi^{*} = \argmax_{\pi} \mathbb{E}_{\rho}[ \sum^{\infty}_{t=0} \gamma^tr(s_t,a_t) |a_t{\small \sim}\pi(\cdot|s_t), s_{t+1}{\small \sim} \mathcal{P}(\cdot|s_t) ] 
	\end{align}
	
	\subsection{Task Formulation}\label{sec: task formulation}
	We train and test our mixed $H_{\infty}$-RL controller (Fig.~\ref{mixture}) to perform navigation tasks in the simulated environments. The agent's goal is to control the vehicle to a desired position. We first introduce two environments: a simplified toy environment, \textit{TurtleSim}, and the blimp simulator\cite{price2022simulation}.
	
	\subsubsection{Turtle Control Task}
	\label{sec: turtle control task}
\begin{figure*}[t]
		\centering
		\begin{subfigure}[t]{0.5\textwidth}
			\centering
			\includegraphics[width=0.9\linewidth]{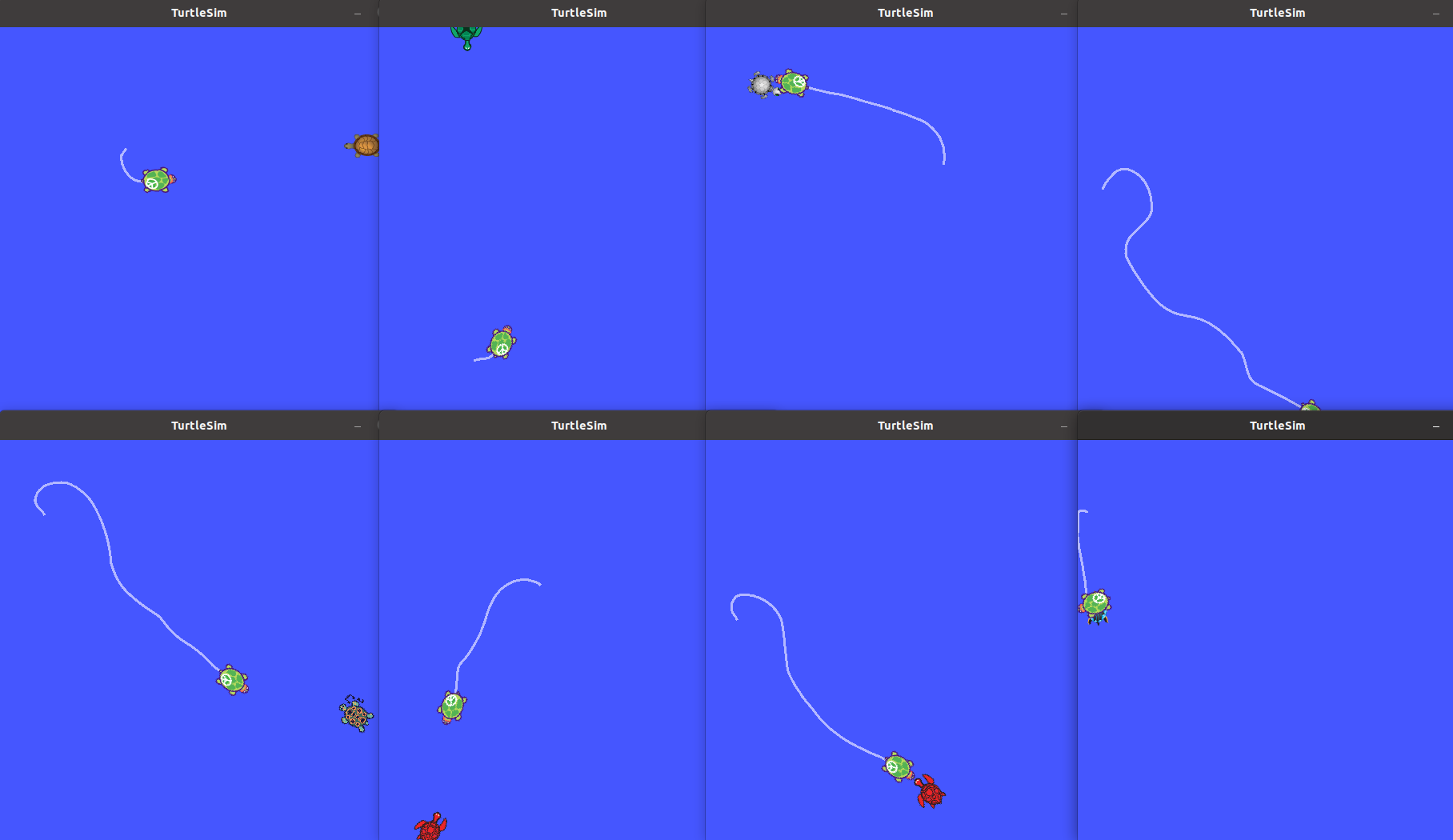}
			\caption{TurtleSim with parallel data collection. The green turtle is the robot, and the target position is represented by a turtle in another color. The white curve displays the position odometry.}
			\label{fig: turtle}
		\end{subfigure}
		\hfill
		\begin{subfigure}[t]{0.45\textwidth}
			\centering
			\includegraphics[width=0.8\linewidth]{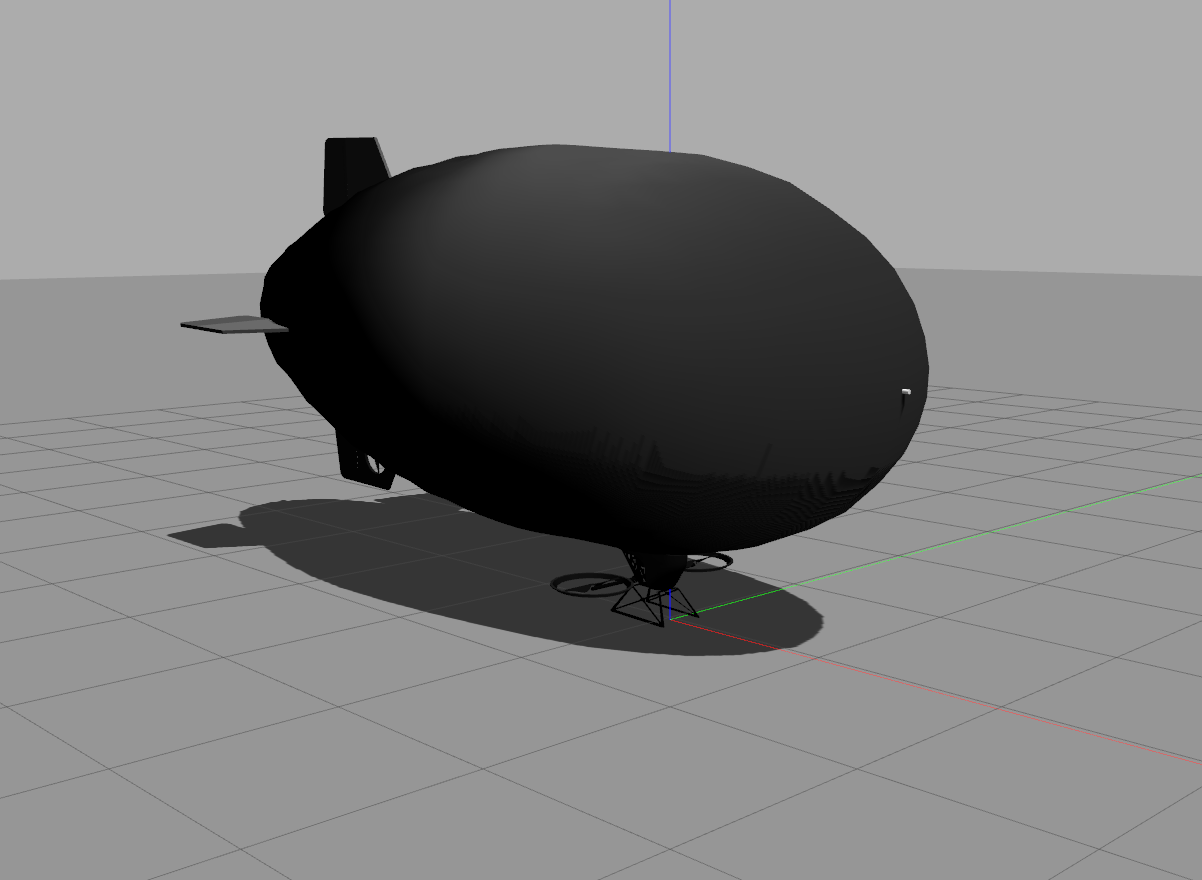}
			\caption{The blimp simulator is implemented in ROS/Gazebo framework. It provides high-fidelity fluid dynamics, and supports software-in-the-loop simulation (SITL) \cite{price2022simulation}.}
			\label{fig: blimp}
		\end{subfigure}
		\caption{Simulation Environments}
	\end{figure*}

	Due to the similarity to the blimp control problem, we introduce it for the ablation study. As shown in Fig.~\ref{fig: turtle}, the agent observes the state in every time step and controls the robot turtle to a stationary target position. Both robot and target positions spawn randomly in every new episode. We formulate the problem by an MDP with the following state and action space,
	
	\begin{itemize}
		\item state space: $s_t=(s_{\theta},s_{l} , u_v, u_{\omega}, q)_t \in [0,1]$,
		\item action space: $a_t=(a_v, a_{\omega})_t \in [-1,1]$,
	\end{itemize}
	where all states are in the range $[0,1]$, and the scaled state $(s_{\theta}, s_{l})$ are the relative yaw angle $\theta$ and relative distance $l$, augmented with the mixing factor $q\in [0,1]$, and the base control  $({u_v}, {u_{\omega}})$ corresponds to thrust and yaw velocity command and share the same command channel with the agent's actions $(a_v, a_{\omega})$. Then the navigation task can be formulated by the reward function, 
	\setlength\abovedisplayskip{0pt} \setlength\belowdisplayskip{0pt}
	\begin{align}
		&r_t = \begin{bmatrix}
			w_{success} & w_{track}  \\
		\end{bmatrix}   \begin{bmatrix}
			r_{success,t}    \\
			r_{track,t}    \\
		\end{bmatrix} \\
		&r_{success,t} = 1 \; if \; |l_t| \leq \epsilon \; else \; 0,\\
		&r_{track,t} =-|l_t|, 
	\end{align}
	where by default $w_{success}=500$, $w_{track}=0.1$, and $\epsilon=0.1$. The environment resets itself when the success reward is obtained. 
	
	\subsubsection{Blimp Control Task}
	\label{sec: blimp control task}
	
	Similar to the turtle control task, the goal of the RL agent is to navigate the robotic blimp (Fig.~\ref{fig: blimp}) to a virtual position target. The state and action space are specified as follows,
	
	\begin{itemize}
		\item state space: $s_t=(s_{z}, s_{l}, s_{\theta}, u_{\zeta}, u_{\eta}, u_{\epsilon}, u_{\delta}, q )_t $
		\item action space: $a_t=( a_{\zeta}, a_{\eta}, a_{\epsilon}, a_{\delta})_t$
	\end{itemize}
	where all states are scaled in the range $[0,1]$, and the scaled state $(s_{z}, s_{l}, s_{\theta})$ are the relative altitude $z$, relative distance $l$, relative yaw angle $\theta$, augmented with the mixing factor $q\in [0,1]$ and the base control command $(u_{\zeta}, u_{\eta}, u_{\epsilon}, u_{\delta})$ corresponding to the control of rudder deflection, elevator deflection, the servo thrust angle, and the thrust magnitude. The actions $(a_{\zeta}, a_{\eta}, a_{\epsilon}, a_{\delta})$ corresponds to the same command channels. Note that the action dynamics are coupled; for example, one can ascend by an elevator when moving forward or directly thrusting upward through the thrust vector.
	
	In this context, there are two major differences from the prior work\cite{Liu.20220310}. First is the usage of the reverse thrust. Descending a blimp is challenging since the blimp's heading velocity is usually slow, and, consequently, the altitude descent velocity from the elevator is also slow. This can cause significant altitude tracking errors. And therefore, even though reverse thrusting is generally less efficient, it helps the blimp descend much faster when lacking the heading velocity. 
	
	Another difference is that we trigger the next target waypoint only when the total distance to the target is less than a threshold of 5 meters instead of the planar distance. This requires much higher efficiency over the altitude control and poses a more significant challenge for control allocation as there are diverse ways to achieve it, e.g., elevator or thrust vector. We demonstrate in the experiment that the RL agents fail to find any viable control policy without efficiently using thrust vectoring.
	
	The following reward function formulates the navigation task,
	\begin{align}
		&r_t = \begin{bmatrix}
			w_{success} & w_{track} & w_{penalty}  \\
		\end{bmatrix}   \begin{bmatrix}
			r_{success,t}    \\
			r_{track,t}    \\
			r_{penalty,t}    \\
		\end{bmatrix} \\
		&r_{success,t} = 1 \; if \; |l_t| \leq \epsilon \; else \; 0,\\
		&r_{track,t} =-w_{z}|z_t| -w_{l}|l_t| -w_{\theta}|\theta|, \\
		&r_{penalty,t} = \Delta(a,u), 
	\end{align}
	where the default value of the task weight is $(w_{success}, w_{track}, w_{penalty})=(500,1,10)$, the tracking reward weight $(w_z, w_l, w_{\theta})=(2,5,2)$ and $\epsilon=5$[m]. The term $\Delta(a,u)$ penalizes when the action deviates too much from the base control to encourage the synergy between the agent and the controller. In practice, we found out that without this ad-hoc penalty, RL agents fail to find any viable control policy. At each time step, we initialize $\Delta(a,u)=0$, and then accumulate it if any of the conditions are triggered,
	\begin{itemize}
		\item  $\Delta(a,u)\pluseq-0.5$, if $a_{\zeta} u_{\zeta} <0$ and $|a_{\zeta}-u_{\zeta}|>0.4$. 
		\item  $\Delta(a,u)\pluseq-0.5$, if $a_{\eta} u_{\eta} < 0$ and $|a_{\eta}-u_{\eta}|>0.4$. 
		\item  $\Delta(a,u)\pluseq-0.5$, if $a_{\epsilon} a_{\delta}>0$ .
		\item  $\Delta(a,u)\pluseq1$, if $a_{\epsilon} u_{\epsilon}>0$. 
		\item  $\Delta(a,u)\pluseq-0.5$, if $u_{\eta} = -1$, $u_{\epsilon} = 0.5$ and $a_{\epsilon} > 0.7$. 
	\end{itemize}  
	To avoid misuse of the reverse thrust, the third condition penalizes when the agent commands the thrust vector to tilt backward $a_{\epsilon}>0$ and the thrust $a_{\delta}$ to be positive and vice versa. Similarly, the last condition penalizes when the controller commands the thrust vector to tilt backward at its maximum $u_{\epsilon}=0.5$, but the RL agent tilts the thrust vector even more, which leads to inefficient reverse thrusting. Lastly, all other conditions penalize when the action commands between the agent and controller are too different.

	\subsection{$H_{\infty}$ Robust Control}
	\label{sec: hinf robust control}
	
	\begin{figure}[t]
		\centering
		\begin{tikzpicture}[auto, node distance=1.2cm,>=latex']		
			\node [input, name=input] {};
			\node [sum, right of=input] (sum) {};
			\node [sknot, right of=sum] (knot1){};
			\node [block, right of=knot1] (controller) {$K$};
			\node [sum, right of=controller] (sum2){};
			\node [sknot, right of=sum2] (knot2){};
			\node [sknot, above of=sum2] (du){};
			\node [block, right of=knot2] (system) {$G$};
			\node [sum, right of=system] (sum3){};
			\node [sknot, right of=sum3] (knot3){};
			\node [sknot, right of=knot3] (knot4){};	
			\node [sknot, below of=sum3] (dy){};
			\draw [->] (controller) -- node[name=u] {} (sum2);
			\draw [-] (sum2) -- node[name=u] {} (knot2);
			\draw [->] (knot2) -- node[name=u2] {$u$}(system);
			\node [soutput, right of=knot4] (output) {};
			\node [block, below of=u] (measurements) {$Sensor$};	
			\node [vblock, above of=knot3] (weight s) {$W_S$};
			\node [vblock, above of=weight s] (weight ks) {$W_{KS}$};
			\node [vblock, above of=weight ks] (weight t) {$W_T$};
			\node [output, right of=weight t] (zz1) {};
			\node [output, right of=weight ks] (zz2) {};
			\node [output, right of=weight s] (zz3) {};	
			\draw [draw,->] (input) -- node[name=w] {$w$} (sum);
			\draw [-] (sum) -- node[name=e] {$e$} (knot1);
			\draw [->] (knot1) -- node[name=e2] {} (controller);
			\draw [->] (du) -- node[name=du, pos=0]{$du$}(sum2);
			\draw [->] (system) -- node[name=y, pos=0.5] {}(sum3);
			\draw [-] (sum3) -- node[name=y] {}(knot3);
			\draw [->] (dy) -- node[name=dy, pos=0]{$dy$}(sum3);
			\draw [->] (knot3) -- node[name=y]{$y$}(output);
			\draw [->] (knot4) |- (measurements);
			\draw [->] (knot3) -| (weight s);
			\draw [->] (knot2) |- (weight ks);
			\draw [->] (knot1) |- (weight t);
			\draw [->] (measurements) -| node[pos=0.95] {$-$} node[pos=0.7] {$y_m$}(sum); 			
			\draw [->] (weight t) -- node[name=z1] {$z_1$} (zz1);
			\draw [->] (weight ks) -- node[name=z2] {$z_2$} (zz2);
			\draw [->] (weight s) -- node[name=z2] {$z_3$} (zz3);
		\end{tikzpicture}
		\caption{\textit{$H_{\infty}$} control framework. The goal is to design controller $K$ such that the robot G can be stabilized and input and output disturbance ($du$ and $dy$) can be rejected.}
		\label{fig: hinf}
	\end{figure}
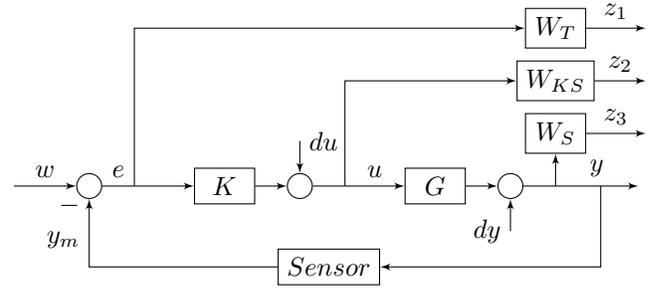

	This framework can be illustrated by the feedback control loop in Fig.~\ref{fig: hinf}, where K is the controller, G is our robot, and the state is directly observed from the sensor without any estimator. Given the tracking signal $w$ and feedback observation $y_m$, The goal is to design the controller such that the tracking error $e=w-y_m$ can be reduced over time subject to input and output disturbance $du$, $dy$. Note that in the hybrid control Fig.~\ref{mixture}, the agent action can be viewed as part of $du$. The weighting filters have the following forms,
	\begin{subequations}
		\label{weighting}
		\begin{empheq}[left=\empheqlbrace]{align}
			&W_{T}(s) = \frac{1}{T_{min}} \cdot \frac{s+\omega_{zt}}{s+\omega_{nt}},  \\
			&W_{KS}(s) = \frac{1}{KS_{min}} \cdot \frac{s+\omega_{zks}}{s+\omega_{nks}} , \\
			&W_{S}(s) = \frac{1}{S_{max}} \cdot \frac{s+\omega_{zs}}{s+\omega_{ns}} , 
		\end{empheq}
	\end{subequations}
	where $\omega$ is the cut-off frequency of each filter. The weight filter parameters in our experiments are presented in Table.~\ref{weighting parameters}. Then the controller K can be solved by satisfying the following constraint, 
	\begin{equation}
		\label{eqn: hinf}	
		\norm{
			\begin{matrix}
				W_S \cdot S   \\
				W_{KS} \cdot KS   \\
				W_T \cdot T  \\
		\end{matrix}}_{\infty} \leq 1
	\end{equation}  
	In practice, we design the weight filters manually and then solve the controller K in MATLAB \cite{MATLAB:2022}. 
	Let $DU$ and $U$ be the Laplace transform of the unknown input disturbance $du$ and control command $u$, then system response can be formulated as follows,
	\begin{equation}
		\label{eqn: uncertainty}	
		G \cdot (I+\Delta \cdot W_T) \cdot U = G \cdot (U+DU),
	\end{equation}  
	by treating the input disturbance as part of model uncertainty, where $I$ is the identity matrix and $\Delta=\Delta(s)$ is the uncertainty matrix with $\norm{\Delta}_{\infty} \leq 1$. After factoring out $G$, we can derive the following relation by matrix sub-multiplicative,
	\begin{align}
		\norm{DU}_2 &= \norm{\Delta  \cdot W_T \cdot U}_2 \\
		&\leq \norm{\Delta}_2 \cdot \norm{W_T}_2 \cdot \norm{U}_2\\
		&\leq \norm{W_T}_2 \cdot \norm{U}_2 \label{eqn: dueq}	
	\end{align}
	Since matrix $W_T$ is diagonal and only consists of identical values, we can consider the $i$-th row of \eqref{eqn: dueq}:
	\begin{equation}
		\label{eqn: dueq2}	
		\norm{DU_{i}}_2 \leq \norm{W_{T,i}}_2 \cdot \norm{U_i}_2.
	\end{equation}  
	
	Note that \eqref{eqn: dueq2} is sufficient but not necessary for \eqref{eqn: dueq}, which means that \eqref{eqn: dueq2} is a more strict condition to determine the upper bound of $\norm{DU}_2$. And by \textit{Parseual's theorem}, we have two identities for \eqref{eqn: dueq2}:
	\begin{subequations}
		\label{eqn: ts}
		\begin{empheq}[left=\empheqlbrace]{align}
			&\norm{DU_{i}(j \omega)}_2 = \norm{du_{i}(t)}_2 \\
			&\norm{U_i(j \omega)}_2 = \norm{u_i(t)}_2 
		\end{empheq}
	\end{subequations}
	Finally, we can derive a conservative upper bound for the plant input disturbance from \eqref{eqn: dueq2} and \eqref{eqn: ts}, i.e., $\norm{du_{i}}_2 \leq \norm{W_{T,i}}_2 \norm{u_i}_2$, such that the $H_{\infty}$-controller stabilizes the plant G. We derive the theoretical upper bound for $\norm{du_{i}}_{2}$ in both simulators as shown in the Table.~\ref{tab: max du}, assuming that the controller commands a step input $u_{i}=1$ when $t \geq 0$. 
	
	\begin{table}[htbp]
		\setlength{\tabcolsep}{4pt}
		\begin{center}
			\begin{tabular}{|c|c|c|c|c|}
				\hline
				\textbf{Simulator} & $\norm{W_{T,i}}_2(j \omega)$ & $\norm{u_i}_{2}$ & Maximum $\norm{du_{i}}_{2}$ & $\omega$ $[rad/s]$ \\
				\hline
				\textit{TurtuleSim} & 1.84 & $\sqrt{t}$ & $1.84 \sqrt{t}$ & 100  \\
				\hline
				Blimp & 33.34 & $\sqrt{t}$ & $33.34 \sqrt{t}$ & 10 \\
				\hline
			\end{tabular}
		\end{center}
		\caption{Maximum allowed input disturbance. Depending on the sampling frequency $\omega$, we increase $\norm{W_T}_2$ of the blimp simulator for more robustness and less of TurtleSim for more performance. Symbol $t$ denotes the time duration the controller sending step input $u_{i}=1$.}
		\label{tab: max du}
	\end{table}
	
	Now consider the mixed command in (\ref{eqn: mixture}). As long as the following relation is satisfied, then the process will remain stable.
	\begin{align}
		\norm{(1-q(t)) \cdot (u_i(t)-a_i(t))}_2 \leq \norm{W_{T,i}(j \omega)}_2 \cdot \norm{u_i(t)}_2
		\label{eqn: constraint}
	\end{align}
	Intuitively, the RHS is the upper bound of the input disturbance  the base control $u$ can reject. If we choose the weighting parameter $q$ as a constant through time, we can further reduce \eqref{eqn: constraint} to :
	\begin{align}
		q \geq 1-\frac{\norm{W_{T,i}(j \omega)}_2 \cdot \norm{u_i(t)}_2}{\norm{(u_i(t)-a_i(t))}_2} 
		\label{eqn: constraint2}
	\end{align}
	Then according to Table.~\ref{tab: max du} and the constraint (\ref{eqn: constraint2}) and assuming \textit{average case} when agent actions are uniform random and have an average zero, i.e., $\mathbb{E}\left[ a \right]= 0$, we can select any distribution for the mixing factor $q$ as long as the mean of the distribution is positive, i.e., $\mathbb{E}\left[ q \right]\geq 0$. Or assuming \textit{worst-case} scenario when we have an adversarial agent, i.e., $\mathbb{E}\left[ a \right]= -u$, then when $\mathbb{E}\left[ q \right]\geq 0.08$ for TurtleSim or  $\mathbb{E}\left[ q \right]\geq 0$ for the blimp simulator, the process will remain stable. 
	
	However, because our plant model is likely imperfect and considering other disturbance and noise that is not modeled, the allowed maximum input disturbance will be less than the estimation. In practice, our conservative choice of $\mathbb{E}\left[ q \right]=0.5$ for the TurtleSim and $\mathbb{E}\left[ q \right]=0.3$ for the blimp simulator seem to work well. 
	
	\begin{table}[htbp]
		\begin{center}
			\begin{tabular}{|c|c|c|c|c|}
				\hline
				\textbf{Parameters} & \textbf{\textit{turtle}} & \textbf{\textit{blimp, yaw}} & \textbf{\textit{blimp, as}} & \textbf{\textit{blimp, ds}}  \\
				\hline
				$T_{max}$ & $\sqrt{5}$ & $\sqrt{2.2}$ & $\sqrt{2}$ & $\sqrt{2}$ \\
				\hline
				$T_{min}$ & $0.01$ & $0.01$ & $0.01$ & $0.01$ \\
				\hline
				$KS_{max}$ & $10$ & $0.8$ & $0.5$ & $0.5$ \\
				\hline
				$KS_{min}$ & $0.01$ & $0.01$ & $0.01$ & $0.01$ \\
				\hline
				$S_{max}$ & $2\sqrt{2}$ & $\sqrt{2}$ & $\sqrt{2}$ & $\sqrt{2}$ \\
				\hline
				$S_{min}$ & $0.001$ & $0.01$ & $0.01$ & $0.01$ \\
				\hline
				$\omega_{ns}$ & $0.00625\sqrt{2}$ & $0.001\sqrt{2}$ & $0.001\sqrt{2}$ & $0.001\sqrt{2}$ \\
				\hline
				$\omega_{zt}$ & $25$ & $0.2$ & $0.2$ & $0.2$ \\
				\hline
				$\omega_{nt}$ & $2500\sqrt{5}$ & $20\sqrt{2.2}$ & $20\sqrt{2}$ & $20\sqrt{2}$ \\
				\hline
				$\omega_{zks}$ & $200$ & $0.8$ & $0.5$ & $0.5$ \\
				\hline
				$\omega_{nks}$ & $2 \cdot 10^5$ & $64$ & $25$ & $25$ \\
				\hline
				$\omega_{zs}$ & $25$ & $0.2$ & $0.2$ & $0.2$  \\
				\hline
				\multicolumn{5}{p{8cm}}{}
			\end{tabular}
			\caption{$H_{\infty}$ control weighting filter parameters. Note that column "turtle" denotes the $H_{\infty}$-controller in \textit{TurtleSim}, and "blimp, yaw/as/ds" denote the yaw, ascend, and descend motion, respectively, in the blimp simulator.}  
			\label{weighting parameters}
		\end{center}
	\end{table} 
	
	
	\subsection{Robust Hybrid $H_{\infty}$-RL in TurtleSim}
	\label{sec: robust hinf-rl in turtlesim}
	Given the controller command $u=[u_v, u_{\omega}]^T$, reference $w=[0,0]^T$, the state vector $x=[l, \theta]^T$, and plant output $y = [l, \theta]^T$, the kinematics of the turtle can be represented by the following state space model,
	\begin{align}
		\label{eqn: turtle ss}
		&A_{turtle} = \begin{bmatrix}
			0 & 0     \\
			0 & 0    \\
		\end{bmatrix},
		B_{turtle} = \begin{bmatrix}
			-1      \\
			1       \\
		\end{bmatrix},  \\ 
		&C_{turtle} = \begin{bmatrix}
			1 & 0     \\
			0 & 1    \\
		\end{bmatrix},
		D_{turtle} = \begin{bmatrix}
			0    \\
			0   \\
		\end{bmatrix}.
	\end{align} 
	
\begin{figure}[h]
		\centering
		\begin{tikzpicture}[auto, node distance=1.6cm,>=latex']		
			\node [input, name=input] {};
			\node [knot, right of=input] (node1) {};
			\node [block, right of=node1] (plant) {$P$};
			\node [knot, right of=plant] (node2) {};
			\node [output, right of=node2] (output){};
			\node [block, below of=plant, node distance=1cm] (controller){$K$};
			\draw [->] ([yshift=0.1cm]input) -- node[name=w, pos=0.2] {$w$} (plant.170);
			\draw [->] (plant.10) -- node[name=z, pos=0.8] {$z$} ([yshift=0.1cm]output);
			\draw [->] (plant.350) -- ([yshift=-0.08cm]node2 |- plant) |- node[name=e, pos=0.25]{$e$} (controller);
			\draw [->] (controller) -- (node1 |- controller) |- node[name=u, pos=0.25]{$u$} (plant.190);
		\end{tikzpicture}
		\caption{$H_{\infty}$ controller compact form}
		\label{fig: hinf2}
	\end{figure}
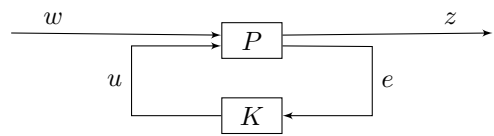

	Recall that $l$ denotes the relative distance between the turtle and its target, and $\theta$ represents the heading angle difference to the target. The kinematic model of the turtle, i.e., the plant $G$, including the disturbances $du$ and $dy$, sensor measurement, and the weighting filters $W$, yields the augmented plant $P$ (Fig.~\ref{fig: hinf2}). The augmented plant $P$ is stabilized by a controller $K$, which is obtained by applying the $H_{\infty}$ design method given the constraint (\ref{eqn: hinf}) and Table.~\ref{weighting parameters}. 
	
	Now consider the hybrid scenario, and the control command has the following form,
	\begin{equation}
		a_{mixed} = \begin{bmatrix}
			(1-q)a_v + qu_v  &  (1-q)a_{\omega} + qu_{\omega} \\
		\end{bmatrix}^T
		\label{eqn: actsturtle}     
	\end{equation} 
	where $a \sim \pi(\cdot|s)$. The mixing factor is sampled randomly from the uniform random distribution at each time step, i.e., $q\sim\mathcal{U} (0,1)$, which satisfies the constraint in (\ref{eqn: constraint2}). Note that it is important for the agent to observe the entire interval of $q\in[0,1]$ during training so that the agent will be able to generalize to any arbitrary $q$ in the testing phase.
	
	\subsection{Robust Hybrid $H_{\infty}$-RL in the Blimp Simulator}
	\label{sec: robust hinf-rl in bimpsim}
	
	Since one compact \textit{MIMO (multiple-input and multiple-output)}-controller using the $H_{\infty}$-method is hard to derive, we split the entire blimp dynamic into two linear sub-systems: the yaw motion and the others. 
	
	The yaw dynamic is modeled as $(A,B,C,D)_{yaw}=(0,-20,1,0)$, with the state $x_{yaw} = \theta$, $y_{yaw} = \theta$, $u_{yaw} = u_{\zeta}$ and $w_{yaw} = \theta_{ref}$, where $\theta$ is the heading angle of the blimp w.r.t the world frame. Furthermore, we model the time delay by the second-order \textit{Padé approximation} with a dead time $T=0.65$,
	\begin{equation}
		e^{-Ts} \approx \frac{2-Ts}{2+Ts}
		\label{eqn: pade}     
	\end{equation}  
	The rest of the dynamics is required for the velocity and altitude control.
	Since the thrusting angle introduces non-linearity, we linearize at two trim points and design two $H_{\infty}$-controllers for the ascending and descending motions, respectively. In both modes, we have the same states, plant outputs, and control commands, i.e., $x_{as/ds}=[-l, z]^T$, $y_{as/ds}=[l, z]^T$, $u_{as/ds}=[ u_{\eta}, u_{\epsilon}, u_{\delta} ]^T$, and $w_{as/ds}=[0,0]^T$.
	
	Recall that $l$ denotes the relative distance, and $z$ denotes the relative altitude. Then we have the ascending dynamics, 
	\begin{align}
		\label{eqn: sas}
		&A_{ascend} = \begin{bmatrix}
			0 & 0     \\
			0 & 0    \\
		\end{bmatrix},   
		B_{ascend} = \begin{bmatrix}
			0 & -5 & 9.8     \\
			-0.4 & -0.77 & 0    \\
		\end{bmatrix}   \\
		&C_{ascend} = \begin{bmatrix}
			-1 & 0     \\
			0 & 1    \\
		\end{bmatrix},   
		D_{ascend} = \begin{bmatrix}
			0 & 0 & 0     \\
			0 & 0 & 0    \\
		\end{bmatrix} 
	\end{align}   
	and descending dynamics,
	\begin{align}
		\label{eqn: sds}
		&A_{descend} = \begin{bmatrix}
			0 & 0     \\
			0 & 0    \\
		\end{bmatrix},   
		B_{descend} = \begin{bmatrix}
			0 & 5 & -9.8     \\
			-0.4 & 0.77 & 0    \\
		\end{bmatrix},   \\
		&C_{descend} = \begin{bmatrix}
			-1 & 0     \\
			0 & 1    \\
		\end{bmatrix},   
		D_{descend} = \begin{bmatrix}
			0 & 0 & 0     \\
			0 & 0 & 0    \\
		\end{bmatrix}. 
	\end{align}  
	As we are applying two linear controllers to two highly non-linear dynamics, we further restrict the controller commands by heuristics to assure that the blimp works near the linearization points, i.e., $u_{\epsilon} \in [-1,-0.5]$ and $u_{\delta} \in [0.4,0.6]$ in ascending mode while $u_{\epsilon} \in [0.5,1]$ and $u_{\delta} \in [-0.6,-0.4]$ in descending mode. 
	
	Now, we obtain in total three $H_{\infty}$ controllers by applying the $H_{\infty}$ design method via (\ref{eqn: hinf}) and the weighting filter (Table.~\ref{weighting parameters}) for their linearized dynamics. The altitude controller switches the mode at zero relative altitudes, i.e., $z=0$, while the yaw controller remains independent of the altitude control. 
	
	Finally, our hybrid DRRL agent has the action in the format, 
	\begin{equation}
		a_{mixed} = (1-q)\begin{bmatrix}
			a_{\zeta}    \\
			a_{\eta}    \\
			a_{\epsilon}   \\
			a_{\delta}    \\
		\end{bmatrix} + q\begin{bmatrix}
			u_{\zeta}    \\
			u_{\eta}    \\
			u_{\epsilon}   \\
			u_{\delta}    \\
		\end{bmatrix} 
		\label{eqn: actsblimp}     
	\end{equation} 
	During training, we sample $q$ in some distributions, while in the testing phase, the controller can provide $q$ based on the constraint (\ref{eqn: constraint2}) or apply a constant $q$ as we did in this work. We have experimented with different $q$ distributions and empirically found that the mixing factor with any distribution works well if it covers the full range $q\in [0,1]$. More details are displayed in the experiment section (Sec.~\ref{sec: experiment blimp}).
	
	\subsection{Training the Robust DRRL Agent}
	\label{sec: training the robust drrl agent}
	Our networks (Table.~\ref{netsize}) follow the actor-critic architecture, which requires two function approximators, e.g., deep neural networks, for the value estimation, $V_{\theta}(s,a)$ and the policy distribution $\pi_{\phi}(\cdot|s)$. A PPO agent (proximal policy optimization, \cite{schulman2017proximal}), with hyper-parameters in Table.~\ref{tab: ppo parameters}, is employed to optimize our networks' parameters.
	
	\begin{table}[htbp]
		\parbox{.5\linewidth}{
			\begin{tabular}{c|c|c|c|c|c| c|c|c|c|c|}
				\cline{2-11} 
				& \multicolumn{5}{|c|}{Value Network} & \multicolumn{5}{|c|}{Policy Network} \\
				\cline{1-11}
				\multicolumn{1}{|c|}{Simulator} & $o$ & $L$ & $F_1$ & $F_2$ & $v$ & $o$ & $L$ & $F_1$ & $F_2$ & $\mu$ \\
				\hline
				\multicolumn{1}{|c|}{TurtuleSim} & 5 & 64 & 64 & 64 & 1   & 5 & 24 & 24 & 24 & 2   \\
				\hline
				\multicolumn{1}{|c|}{Blimp} & 8 & 196 & 196 & 196 & 1   & 8 & 64 & 64 & 64 & 4   \\
				\hline
			\end{tabular}
		}
		\caption{Network architecture. Notation $L$ denotes the LSTM layer, F is the fully connected layer, and the numbers indicate the layer's size. Following the suggestion of a recent work \cite{Andrychowicz.20200610}, we choose \textit{tanh} as our activation function and initialize the last layer with small weights (e.g., 0.01) to improve the exploration.}
		\label{netsize}
	\end{table}
	
	The hybrid agent collects data by interacting with the $N_{env}$ parallelized environment with randomized mixing factor $q$. A waypoint is sampled randomly in every episode, and it will be triggered when the robot is within a certain distance, e.g., 1[m] for the TurtleSim and 5[m] for the blimp simulator. The environment resets when the waypoint is triggered. We inject noise into the observations to increase the agent's robustness and randomize wind disturbance and buoyancy in every episode.
	
	The transition data are stored in the buffer with size $N_{epoch}$. When the buffer is full, the agent will start learning by querying $L/N_{batch}$ mini-batches and updating $N_{update}$ times for each of them or when the KL threshold $D_{KL}$ is reached. Note that the base control can be considered part of the environment in our hybrid control scenario. The agent's goal is to optimize the total amount of reward considering the base control's decision.
	
	\begin{table}[htbp]
		\begin{center}
			\begin{tabular}{|l|c|c|}
				\hline
				\textbf{Parameters} & \textbf{\textit{TurtleSim}} & \textbf{\textit{Blimp}}  \\
				\hline
				Time Steps per Environment $N_{step}$  & 50000 & 86400 \\
				\hline
				Parallelization $N_{env}$ & 8 &  7\\
				\hline
				Episode Length $L$ & 2000 & 2400 \\
				\hline
				Loop Rate [$Hz$] & 100 & 10 \\
				\hline
				Epoch Length $N_{epoch}$  & 1000 & 1920 \\
				\hline
				Mini-batch Size $N_{batch}$  & 100 & 128 \\
				\hline
				Update per Epoch $N_{update}$  & 20 & 20 \\
				\hline \hline
				Initial Policy Learning Rate $\alpha_{0}$ & $5 \cdot 10^{-5}$ & $5 \cdot 10^{-5}$ \\
				\hline
				Initial Value Learning Rate $\beta_{0}$ & $1 \cdot 10^{-4}$ & $1 \cdot 10^{-4}$  \\
				\hline
				KL Threshold $D_{KL}$ & $\infty$ & 0.03 \\
				\hline
				Discount Factor $\gamma$ & 0.999 & 0.99 \\
				\hline
				GAE Smoothing $\lambda$ & 0.95 & 0.9 \\
				\hline
				Gradient Optimizer & $Adam$ & $Adam$\\
				\hline
			\end{tabular}
		\end{center}
		\caption{PPO hyper-parameters. The learning rate is scheduled by multiplying a constant of less than one every episode until it drops to a minimum of 1e5.}  
		\label{tab: ppo parameters}
	\end{table}

	\section{Experiments and Results}
	\label{sec: Experiments}
	The experiment aims to understand whether increasing the robustness of the base control can enable more performance growth and generate a robust and performant controller. 
	
	\subsection{Experiment Setup}
	We perform all experiments on a single computer (AMD Ryzen Threadripper 3960X, 24x 3.8GHz, NVIDIA GeForce RTX2080 Ti, 11GB). The PPO agent and base controllers, i.e., $H_{\infty}$ and PIDs, are implemented based on Pytorch\cite{paszke2019pytorch} to facilitate vectorized computing. Both TurtleSim and the blimp simulator are implemented based on the ROS and the latter is integrated with the Gazebo SITL simulation. 
	
	We compare our robust $H_{\infty}$-RL controller to the previous PID-RL baseline in the TurtleSim (Sec.~\ref{sec: turtle control task}) and blimp simulator (Sec.~\ref{sec: blimp control task}). 
	\begin{enumerate}
		\item $H_{\infty}$-PPO agent: our proposed approach
		\item PID-PPO agent: previous approach \cite{Liu.20220310}. We re-implemented with a randomized mixing factor $q$ and a servo control so that it has the chance to challenge our more difficult waypoint following task.
	\end{enumerate}
	Note when the mixing factor is $q=1$ we recover the base control, and $q=0$ corresponds to the pure PPO agent. In the training phase, $q$ is sampled randomly from different distributions while fixed in the testing phase.

	\subsection{Performance Metric}
	\label{sec: b score}
	Because the total rewards can be pretty misleading as it only reflects an agent's performance tracking one specific waypoint, and each agent's training reward can differ. We introduce a metric, $b_{score}$, to compare the relative performance between the controllers to replace the reward for our path-following tasks. 
	\begin{equation}
		b_{score}(\pi_i) = 100 \cdot \left ( 1 - \frac{T_{\pi_i}}{\sum_{\pi_j \in \pi}T_{\pi_j}} - \frac{E_{\pi_i}}{\sum_{\pi_j \in \pi}E_{\pi_j}}  \right ),
		\label{eqn: fti}     
	\end{equation} 
	where $T_{\pi_i}$ is the average amount of time for each controller to complete the task and $E_{\pi_i}$ computes the average control effort for each controller. The more time or energy the controller consumes to complete the task, the worse the score relative to other controllers. 
	
	\subsection{TurtleSim}
	\label{sec: experiment turtle}
	We sample a random position target in every new episode to test the proposed hybrid agent. The environment resets when the robot reaches the target position. The experiment terminates when the agent successfully controls the robot to the target position 100 times. The goal of each agent is to trigger the terminal condition with a minimum amount of time and energy in every episode. Table.~\ref{tab: result turtle} is the result of our experiment in TurtleSim, where the energy penalty for each control policy is defined as $E_{\pi} = 0.25 \cdot |\bar{a}_{mix, v}| + 0.75 \cdot |\bar{a}_{mix, \omega}|$, bar denotes the average over the whole trajectory. Each experiment is conducted with ten random seeds. 
	
	Note that because TurtleSim did not have any dynamics, we applied the disturbance $du$ during training to output action to simulate wind, which is formulated as time dependant noise,
	\begin{align}
		\label{eqn: du}
		a_{mixed,t} &\leftarrow a_{mixed, t}+ \delta_{t}  \\
		\delta_t &= \delta_{t-1} + [n_{v}\quad n_{\omega}]^\top
	\end{align}
	where $\delta_t$ is initialized as zero and bounded in $[-1,1]$, and both noises are sampled from standard Gaussian $n_v,n_{\omega}\sim \mathcal{N}(0,1)$. This noise is amplified five times in the testing phase, increasing the bound to $[-5, 5]$.
	
	\begin{table}[htbp]
		\begin{center}
			\begingroup
			\setlength{\tabcolsep}{1.1pt}          
			\renewcommand{\arraystretch}{1.2}     
			
			\begin{tabular}{|c|c|c|c|c|c|c|}
				\hline
				Controller &  $q$  &      $\bar{|a_{v}|}$  &  $\bar{|a_{\omega}|}$ & $T$  & $E$  &  $b_{score}$ \\ 
				\hline  
				PPO                 & 0  &  5.58  & 4.72  & 9916  & 4.93  & 70.63   \\
				\hline
				$H_{\infty}$-PPO              & 0.5 &5.62  & 5.56  &  11709  & 5.58  & 66.05  \\
				\hline
				\textit{PID}-PPO             & 0.5 &  5.88  & 4.96  & 10055  & 5.19  & 69.66    \\
				\hline
				$H_{\infty}$-PPO            & 1  & 7.18  & 6.85  & 11765  & 6.93  & 61.91   \\
				\hline
				\textit{PID}-PPO             & 1  &  7.32  & 5.69  & 12281  & 6.10  & 63.65   \\
				\hline
				
			\end{tabular}
			\endgroup
			
			\caption{Testing in \textit{TurtleSim}. Each row displays the average value of 10 experiments with different seeds.}
			\label{tab: result turtle}
		\end{center}
	\end{table}
	
	Table.~\ref{tab: result turtle} indicates that the PPO agent alone has the best performance and energy efficiency, followed by the PID and then $H_{\infty}$ controller. Therefore, with less controller intervention $q$, the hybrid agent will achieve better performance. Unsurprisingly, the PID controller performs better than the $H_{\infty}$ controller, which trades both the performance and energy efficiency for more robustness against noise and disturbance. Base controller with more robustness allows training with lower mixing factor $q$, but since TurtleSim is relatively simple and allows both base controllers to train and test with arbitrary $q$, the advantage of $H_{\infty}$ is not well reflected in this experiment.

	\begin{table*}[t]
		\begingroup
		\setlength{\tabcolsep}{1.1pt}          
		\renewcommand{\arraystretch}{1.2}     
		\parbox{1\linewidth}{
			\centering
			\begin{tabular}{|c||c|c|c|c|c|c|c|c|c|c||c|c|c|c|c|c||c|c|c|c|c|c|}
				\hline
				controller & \textbf{q} & $a_{\zeta}$ &	$a_{\eta}$&	$a_{\epsilon}$&	$a_{\delta}$   & $T$   &  $L$     &  $E$      & $b_{score}$   & fail   &  
				\textbf{wind}   & $T$   &  $L$     &  $E$      & $b_{score}$   & fail     &
				\textbf{buoyancy}   & $T$   &  $L$     &  $E$      & $b_{score}$   & fail \\ 
				
				\hline  
				PID-PPO & \textbf{0}  &0.83	&0.90	&1	&1    & 3414  & 43.90  & 0.8705 & 84.38      &  7  &  
				\textbf{0} & 2996 &	37.74 &	0.7252 & 86.50  &	4 &
				\textbf{0.93}      & 2420	& 37.89	& 0.4988 &	88.95 &	8 \\
				\hline
				PID-PPO & \textbf{0.5} &0.67	&0.79	&0.99	&0.79   & 4294 & 66.09	& 0.6943 & 81.24	& 5  & 
				\textbf{0.5}    & 4255	& 70.88 & 0.6048 & 81.38 &	5 &
				\textbf{1}    & 2633	 & 55.99 &	0.6284 & 86.64 &	3 \\
				\hline
				PID-PPO & \textbf{1} &0.47	&0.55	&0.94	&0.57      & 2615 & 38.42	& 0.5048 & 88.36	& 4  & 
				\textbf{1}      & 2464 & 33.55 &	0.4927 & 89.15 &	8   &
				\textbf{1.07}      & 5416 &	43.98 &	0.7562 & 39.73 &	6 \\
				\hline
				$H_{\infty}$-PPO &\textbf{0} &0.52	&0.88	&0.57	&1     & 2646 &	35.97 &	 0.8641	 & 87.00	& 0  & 
				\textbf{0}      & 2464 & 35.12 &	0.6813 & 88.28 &	0   &
				\textbf{0.93}      & 3442 &	35.69 &	0.6875 & 85.60 &	0   \\
				\hline
				$H_{\infty}$-PPO & \textbf{0.5} &0.44	&0.63	&0.61	&0.76   & 2484. & 36.10	& 0.6665 &  88.23   & 	0  & 
				\textbf{0.5}    & 2685 &	38.07 &	0.6830 & 87.49 &	0 &
				\textbf{1}        & 2510 &	36.88 &	0.6720 & 88.08 & 	0\\
				\hline
				$H_{\infty}$-PPO & \textbf{1} &0.47	&0.43	&0.71	&0.55     & 3456 & 37.10 & 0.5075 & 86.19	& 0   & 
				\textbf{1}      & 3438 &	35.98 &	0.6739 & 85.65 &	0 &
				\textbf{ 1.07}      & 2633 &	36.60 &	0.6788 & 87.74 &	0\\
				\hline
			\end{tabular}
			\caption{Robustness test. Each experiment is conducted nine times with the coil trajectory. Fail trials are excluded from computing the $b_{score}$ and marked as \textit{fail}.  The energy penalty is defined as $E_{\pi_i} = 0.15 \cdot \bar{|u_{\zeta}|} + 0.05 \cdot \bar{|u_{\eta}|} + 0.1 \cdot (1-\bar{|u_{\epsilon}|}) + 0.7 \cdot \bar{|u_{\delta}|}$, which penalizes majorly on thrusting. The unit for wind is [m/s] and for buoyancy is [\%]. The maximum speed of the simulated blimp is 2 [m/s].}
			\label{tab: robust test}
		}
		\endgroup
	\end{table*} 
	
	\subsection{Blimp Simulator}
	\label{sec: experiment blimp}
	The training method of the hybrid agent is introduced in Sec.~\ref{sec: training the robust drrl agent} for both PID-PPO baseline and our $H_{\infty}$-PPO agent. We conduct an ablation study on the effect of the different sampling distributions for the mixing factor $q$ during training. The testing phase is conducted in the coil trajectory, represented by a sequence of waypoints with three different wind disturbance and buoyancy levels. Each evaluation finishes when the agent completes the designated trajectory five times. The waypoints are triggered when the robot is within 10 meters, and the following waypoints will become active. 
	
	The coil trajectory consists of 15 waypoints with a 50 meters radius. Each waypoint is placed 45 degrees counter-clockwise from the previous one with 3 meters increase in altitude. The coil trajectory poses a great challenge. Due to the shorter planar distance, the controllers must constantly slow down the blimp to prevent overshooting the waypoints, which can incur significant altitude loss.  
	
	Since deviating far from the track compromises the blimp's safety, the position tracking error L is introduced to the $b_{score}$ for the blimp navigation task, i.e.,
	\begin{align}
		b&_{score}(\pi_i) \notag\\ 
		&= 100 \cdot  \left ( 1 - \frac{3T_{\pi_i}}{\sum_{\pi_j \in \pi}T_{\pi_j}}   - \frac{E_{\pi_i}}{\sum_{\pi_j \in \pi}E_{\pi_j}} - \frac{L_{\pi_i}}{\sum_{\pi_j \in \pi}L_{\pi_j}}   \right ) ,
		\label{eqn: fbi}     
	\end{align} 
	where $T_{\pi_i}$ and $E_{\pi_i}$ are the time and energy penalty, and the $L_{\pi_i}$ is the average distance loss computed by the norm of the relative position. 
	
	The robustness test for the agents is displayed in Table~\ref{tab: robust test}. The $H_{\infty}$-PPO outperforms the PID-PPO combination with a significantly higher success rate regardless of the wind or buoyancy condition. Even when the controller has zero intervention $q=0$, the PPO agent trained with $H_{\infty}$ base control performs much better. And the $H_{\infty}$-PPO performs the best when $q=0.5$ without failing to any condition. This shows that our robust residual RL framework can generate a robust, high-performance controller. The explanations can be found in the table as well. 
	
\begin{figure}[h]
		\centering
		\begin{subfigure}[t]{0.24\textwidth}
			\centering
			\includegraphics[width=1\linewidth]{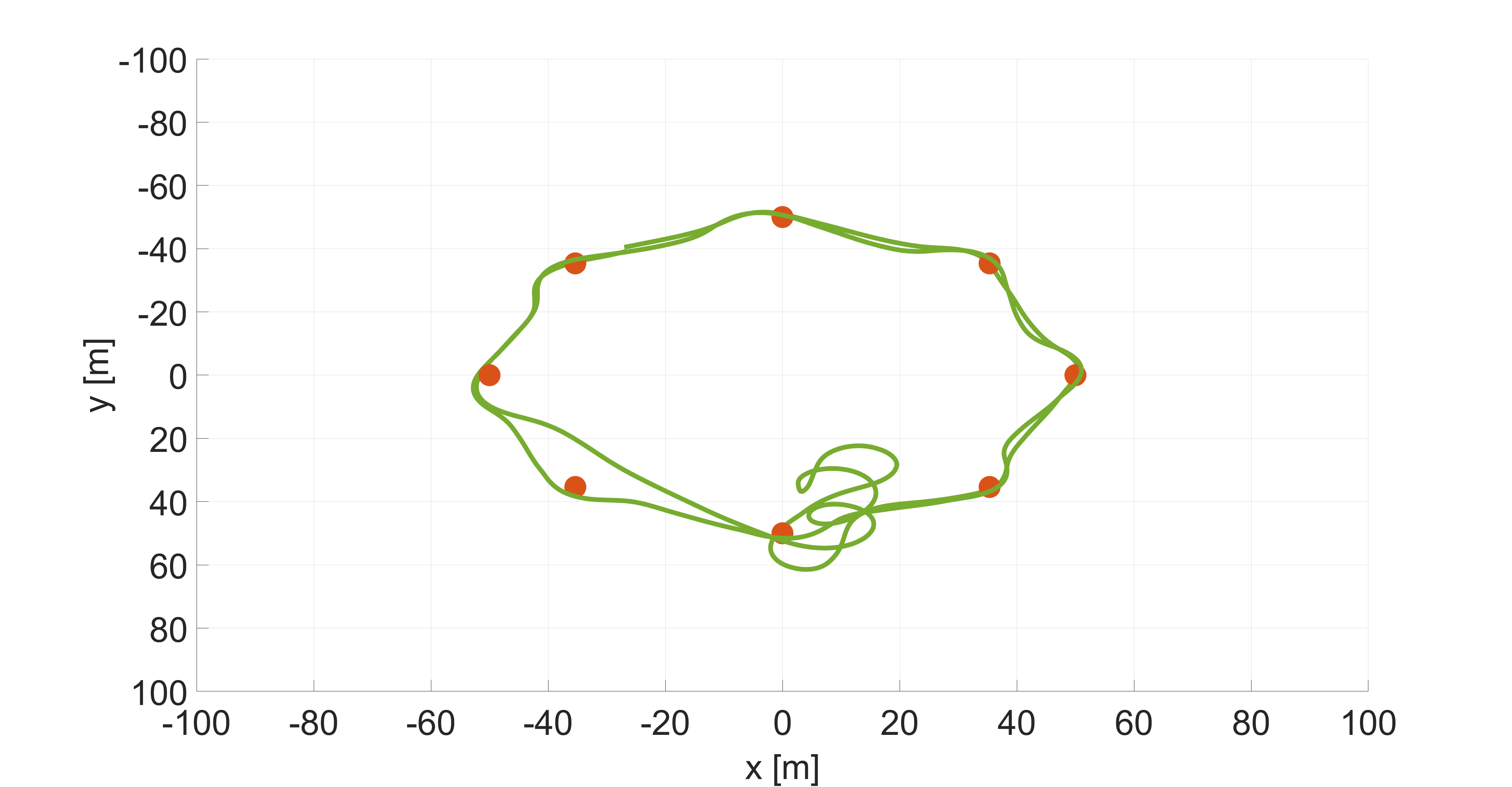}
		\end{subfigure}
		\hspace*{-0.9em}
		\begin{subfigure}[t]{0.24\textwidth}
			\centering
			\includegraphics[width=1\linewidth]{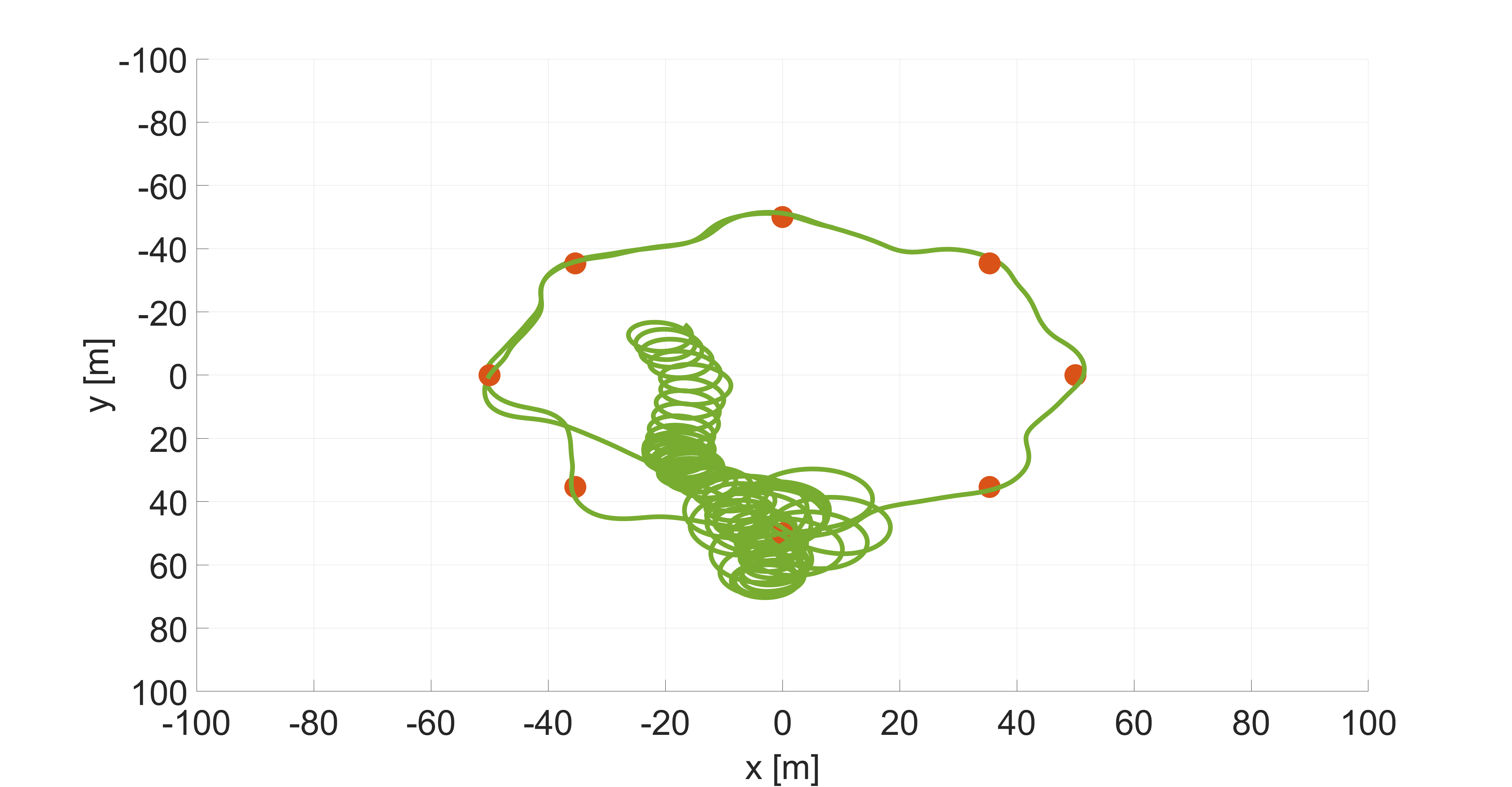}
		\end{subfigure}
		
		\begin{subfigure}[t]{0.24\textwidth}
			\centering
			\includegraphics[width=1\linewidth]{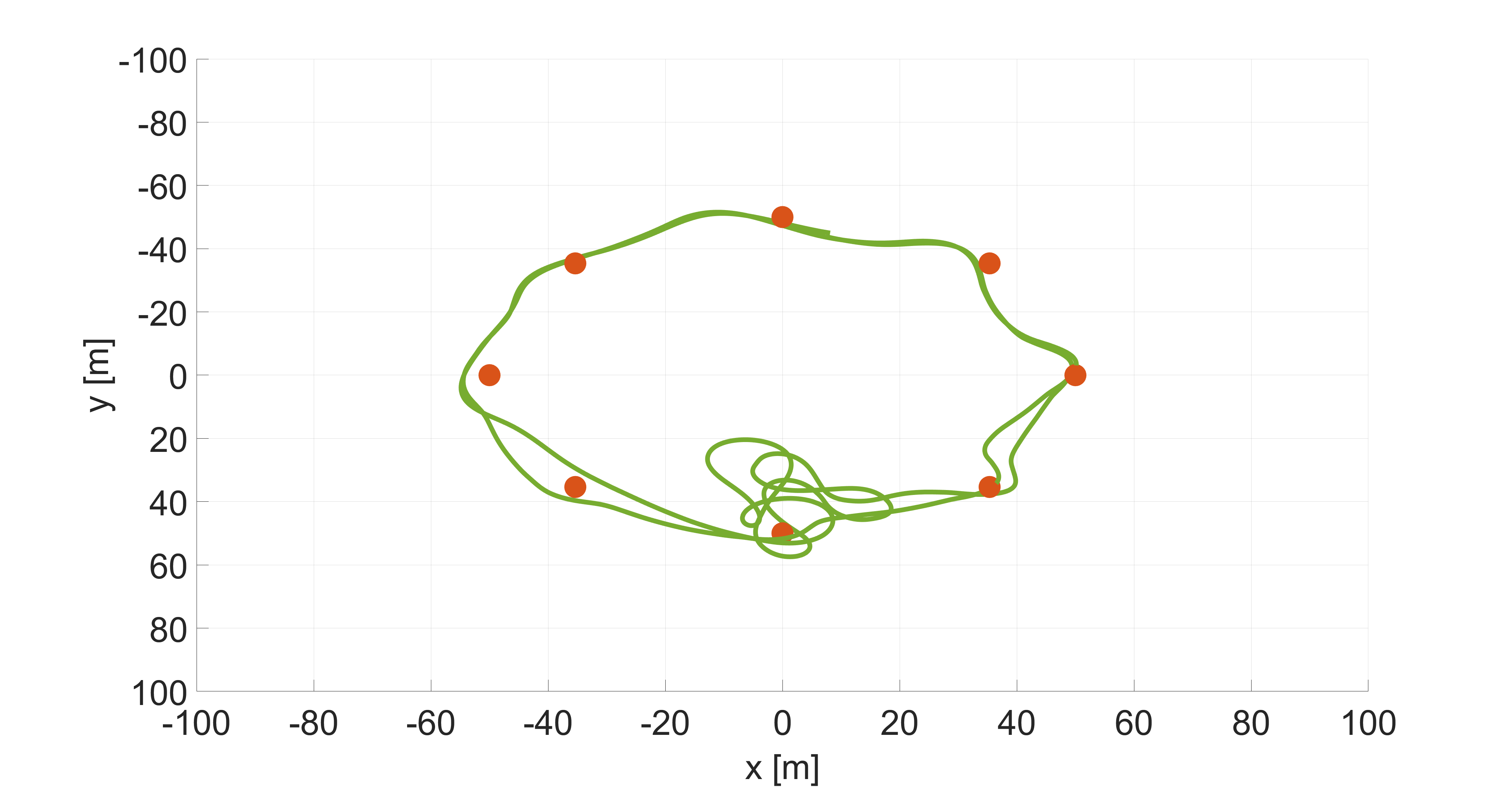}
		\end{subfigure}
		\hspace*{-0.9em}
		\begin{subfigure}[t]{0.24\textwidth}
			\centering
			\includegraphics[width=1\linewidth]{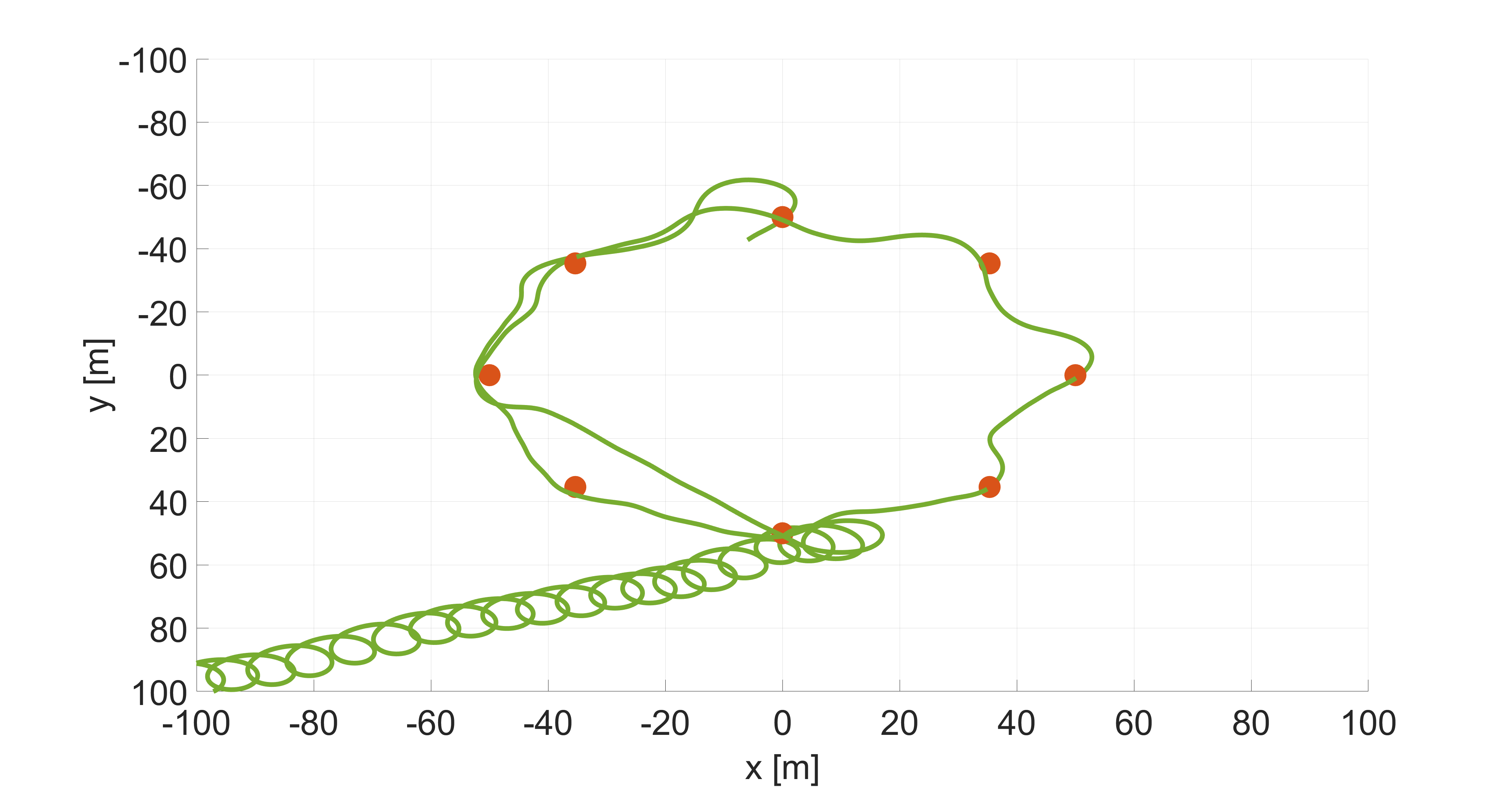}
		\end{subfigure}
		
		\begin{subfigure}[t]{0.24\textwidth}
			\centering
			\includegraphics[width=1\linewidth]{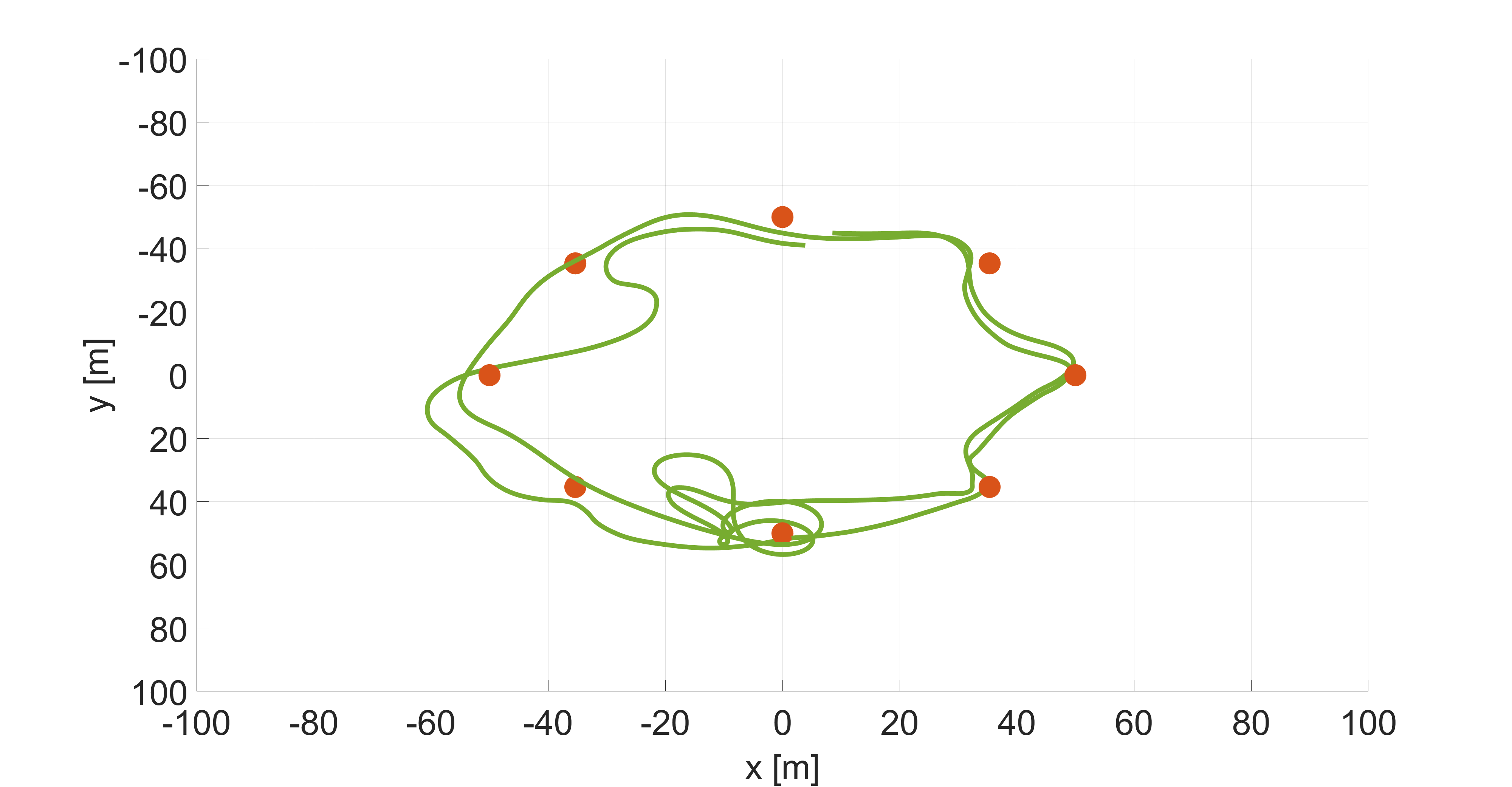}
		\end{subfigure}
		\hspace*{-0.9em}
		\begin{subfigure}[t]{0.24\textwidth}
			\centering
			\includegraphics[width=1\linewidth]{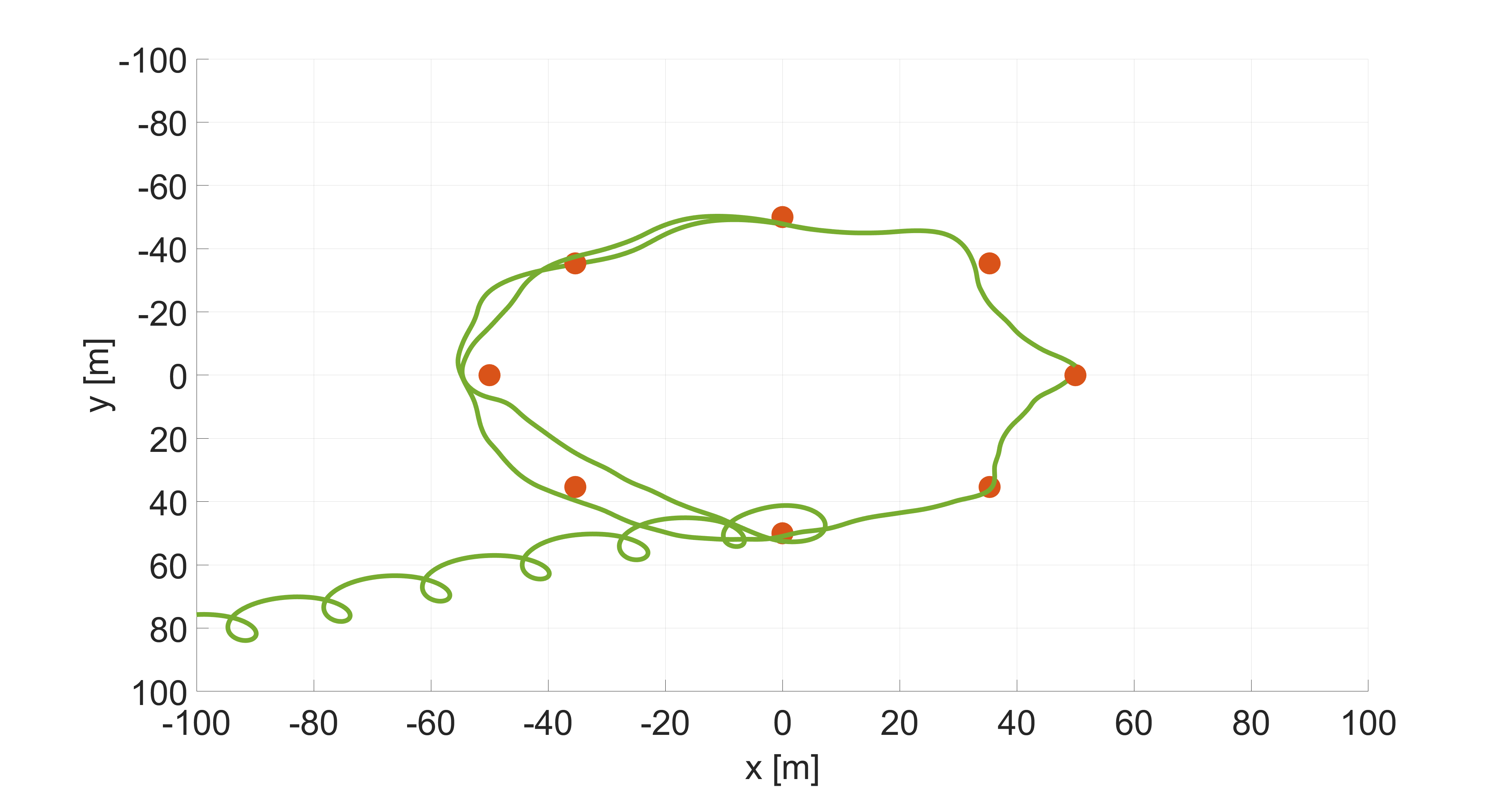}
		\end{subfigure}
		
		\caption{Snippet of the coil trajectories (green curves) and waypoints (red dots). The first and second columns correspond to the $H_{\infty}$-PPO and PID-PPO controllers. Rows correspond to the wind velocities 0, 0.5, and 1 [m/s]. The buoyancy is in nominal condition while the mixing factor $q=0.5$.}
		\label{fig: wind effect}
	\end{figure}

	First, base control robustness is vital to the final performance of the PPO agent. When $q=1$, although $H_{\infty}$ performs worse than PID, its robustness against disturbance secures its success rate. During training, PID can become unstable due to input disturbance from the RL agent, further jeopardizing the PPO training stability while $H_{\infty}$ is less affected. As a result, the PPO trained with PID performs even worse than PID. The wind and buoyancy test reflect the robustness of the controller. The PID-PPO failure rate increases significantly outside the nominal condition in which PID is tuned. The effect of wind can be visualized in Fig.\ref{fig: wind effect}. The wind barely influences $H_{\infty}$-PPO controller while PID-PPO controller is blown away and loses yaw control when descending. Second, the PID controller has relatively poor altitude control even after installing thrust vectoring. Since thrust vectoring introduces high nonlinearity to the system, the PID controller does not benefit much from it. The poor performance in altitude control is reflected in the buoyancy test. 
	
	We conducted an ablation study about the effect of wind disturbance and $q$ distribution during training. Table.~\ref{tab: result wind disturbance} displays the effect of incorporating wind disturbance during training. Regardless of the mixing factor $q$, the $b_{score}$ always decreases when training with the wind. With $q=0$, the performance drops significantly, implying that the wind negatively impacts the agent more than the controller. As a result, we suggest training without any disturbance to encourage aggressive behavior since, under the supervision of the robust controller, the agent no longer needs to behave conservatively.
	
	The effect of mixing factor $q$ during training is displayed in Table.\ref{tab: training mixing factor distribution}. We found the training success rate increases when $q$ becomes larger since the base control is critical in improving the training stability. The distribution with a higher average $q$ is preferred during training as it increases the training success rate. The distribution type is not essential as long as it covers the entire range $q\in[0,1]$. Lastly, as mentioned, a lower average $q$ is desired for improving the control performance in the testing phase.

	\begin{table}[htbp]
		\begingroup
		\setlength{\tabcolsep}{1.1pt}          
		\renewcommand{\arraystretch}{1.2}     
		\parbox{.45\linewidth}{
			\centering
			\begin{tabular}{|c|c|c|c|c|c|}
				\hline
				wind    & $q$   &  $T$  &  $L$   &  $E$  & $b_{score}$ \\ 
				\hline  
				False   & $0.5$ & 7378  & 20.81  & 0.36  & 88.54   \\
				\hline
				True    & $0.5$ & 7727  & 20.66  & 0.44  & 87.58   \\
				\hline
				False   & $0.2$ & 6750  & 19.85  & 0.40  & 88.60   \\
				\hline
				True    & $0.2$ & 7354  & 21.69  & 0.51  & 86.88   \\
				\hline
				False   & $0$   & 6637  & 19.40  & 0.42  & 88.61   \\
				\hline
				True    & $0$   & 8874  & 22.80  & 0.58  & 85.23   \\
				\hline
			\end{tabular}
			\vspace*{5pt}
			\caption{\centering Training with random wind disturbance.}
			\label{tab: result wind disturbance}
		}
		\hspace{-0.3em}
		\parbox{.45\linewidth}{
			\centering
			\begin{tabular}{|c|c|c|c|c|c|}
				\hline
				$q$    &  $T$  &  $L$   &  $E$  &  $b_{score}$ & fail\\ 
				\hline  
				$0.2$                    & 12280  & 26.32    & 0.35  & 85.26  & 2 \\
				\hline
				$\mathcal{U} (0,0.4)$  &  29180  & 20.65    & 0.5  & 77.49  & 1\\
				\hline
				$Beta (2,8)$          & 6921  & 20.02    & 0.39  & 88.58   & 0\\
				\hline
				$0.5$                  & 7523  & 19.69    & 0.45  & 87.84  & 0 \\
				\hline
				$\mathcal{U} (0,1)$  & 6727  & 19.62    & 0.43  & 88.42   & 0\\
				\hline
				$Beta (5,5)$       & 8981  & 19.66   & 0.47  & 86.96    & 0\\
				\hline 			
			\end{tabular}
			\vspace*{5pt}
			\caption{\centering Training with different $q$ distributions.}
			\label{tab: training mixing factor distribution}
		}
		\endgroup
		\caption*{\centering Ablation study with $H_{\infty}$-PPO controller tested in 
			coil trajectory. Every experiment is conducted three times.}	
		\label{tab: ablation study}		
	\end{table}

	\section{Conclusions}
	In this work, we have introduced a $H_{\infty}$-PPO hybrid agent for the blimp control task. We first improve the altitude control efficiency by incorporating the thrust vectoring into the base control and enabling the usage of reverse thrusting. Then, we applied the variable mixing factor, which allows the controller to balance robustness and performance based on the situation. A theoretical lower bound for the mixing factor is derived to guarantee stability. Lastly, we test in the blimp simulator that our robust hybrid agent can outperform the prior PID-PPO combination and demonstrate greater robustness against wind disturbance and buoyancy changes.

	\nocite{*}
	\printbibliography 
	
\end{document}